\documentclass[conference]{IEEEtran}
\IEEEoverridecommandlockouts
\usepackage{cite}
\usepackage{amsmath,amssymb,amsfonts}
\usepackage{algorithmic}
\usepackage{graphicx}
\usepackage{textcomp}
\usepackage{xcolor}
\usepackage{hyperref} 
\usepackage{marvosym}
\usepackage{float}

\def\BibTeX{{\rm B\kern-.05em{\sc i\kern-.025em b}\kern-.08em
    T\kern-.1667em\lower.7ex\hbox{E}\kern-.125emX}}

\usepackage{booktabs}
\usepackage{amsmath}
\usepackage{soul}
\usepackage[utf8]{inputenc}
\usepackage{amsthm}
\usepackage{verbatim}
\usepackage{amssymb}
\usepackage{xcolor}

\begin{document}

\title{ABC-GS: Alignment-Based Controllable Style Transfer for 3D Gaussian Splatting}


\author{\IEEEauthorblockN{Wenjie Liu$^{1,}$\IEEEauthorrefmark{1},
Zhongliang Liu$^{2,}$\IEEEauthorrefmark{1},
Xiaoyan Yang$^{1}$, 
Man Sha$^{3}$ and
Yang Li$^{1,}$\textsuperscript{\Letter}}
\IEEEauthorblockA{$^{1}$School of Computer Science and Technology, East China Normal University, Shanghai, China}
\IEEEauthorblockA{$^{2}$School of Software Engineering, East China Normal University, Shanghai, China}
\IEEEauthorblockA{$^{3}$Shanghai Chinafortune Co., Ltd, Shanghai, China}

\{51265901068,10235101440,51215901035\}@stu.ecnu.edu.cn, shaman@shchinafortune.com, yli@cs.ecnu.edu.cn
\thanks{\IEEEauthorrefmark{1}~Equal contribution. \textsuperscript{\Letter}~Corresponding author.  \protect\\  \protect\hspace{2em}This work is supported by the National Natural Science Foundation of China (62472178, 62376244), and Shanghai Urban Digital Transformation Special Fund Project (202301027).}
\vspace{-3mm}
}

\maketitle

\begin{abstract}
3D scene stylization approaches based on Neural Radiance Fields (NeRF) achieve promising results by optimizing with Nearest Neighbor Feature Matching (NNFM) loss.
However, NNFM loss does not consider global style information.
In addition, the implicit representation of NeRF limits their fine-grained control over the resulting scenes. 
In this paper, we introduce ABC-GS, a novel framework based on 3D Gaussian Splatting to achieve high-quality 3D style transfer.
To this end, a controllable matching stage is designed to achieve precise alignment between scene content and style features through segmentation masks. Moreover, a style transfer loss function based on feature alignment is proposed to ensure that the outcomes of style transfer accurately reflect the global style of the reference image. Furthermore, the original geometric information of the scene is preserved with the depth loss and Gaussian regularization terms. Extensive experiments show that our ABC-GS provides controllability of style transfer and achieves stylization results that are more faithfully aligned with the global style of the chosen artistic reference. Our homepage is available
at \href{https://vpx-ecnu.github.io/ABC-GS-website}{this https url}.

\end{abstract}


\begin{figure*}[h]
  \centering
   \includegraphics[width=1\linewidth]{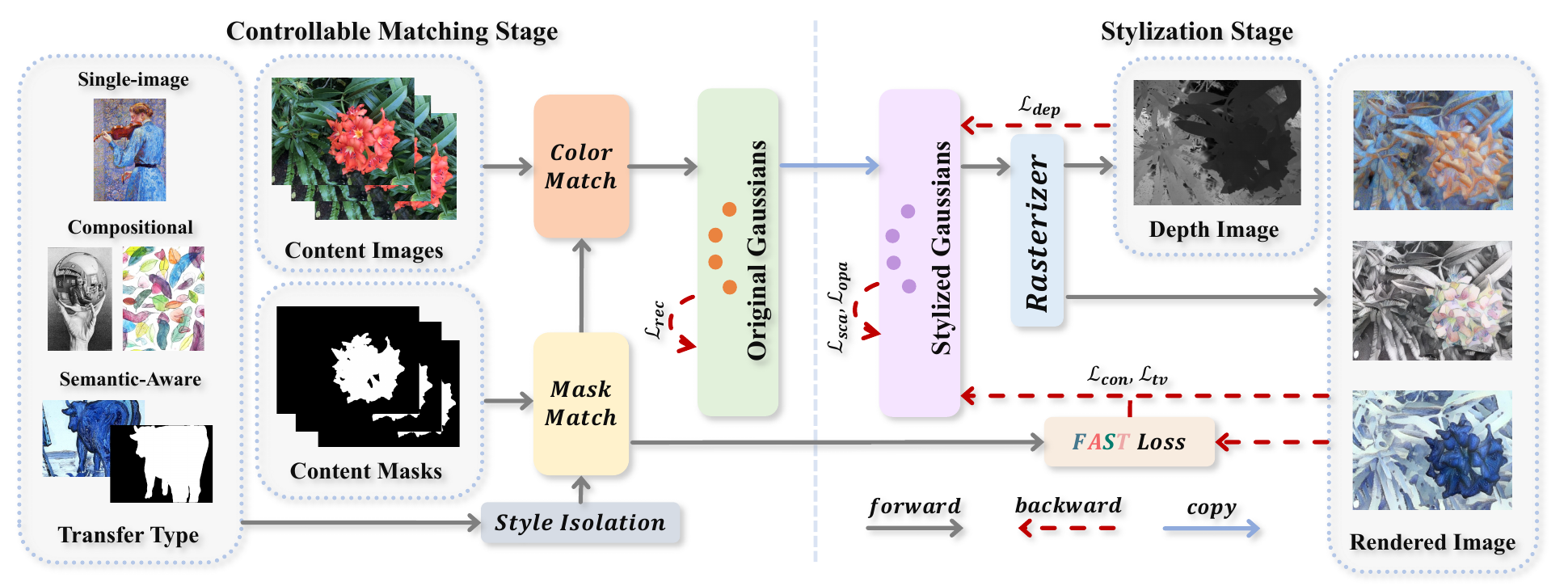}
   \caption{\textbf{ABC-GS Pipeline.} Given a set of content images and content masks, along with style images and style transfer type, our method first achieves a match between style and scene content through mask matching in the controllable matching stage. Subsequently, based on the matching results, color matching that is consistent with the perspective is performed between the content images and the scene Gaussians. In the stylization stage, we use the feature alignment
style transfer loss to optimize the scene and introduce multiple loss terms to maintain the content and geometric
information of the scene.} 
   
   \label{fig:pipeline}
   \vspace{-3mm}
\end{figure*}

\section{Introduction}
In recent years, the demand for 3D stylization technologies has significantly increased, driven by rapid advancements in fields such as virtual reality, augmented reality, and video gaming. {3D stylization enables the transfer of 2D artworks with distinctive visual styles into 3D models, thereby maintaining stylistic coherence. These approaches provide immersive three-dimensional experiences to users.} The introduction of Neural Radiance Fields (NeRF)~\cite{mildenhall2020nerf} has significantly propelled advancements in novel view synthesis, catalyzing developments in 3D stylization. 
Subsequently, a multitude of NeRF-based 3D style transfer techniques have emerged~\cite{huang2022stylizednerf, liu2023stylerf, zhang2022arf, zhang2023ref}. 


Although NeRF-based stylization methods~\cite{zhang2022arf, zhang2024coarf} achieve very promising stylization quality, they usually rely on Nearest Neighbor Feature Matching (NNFM) loss, which introduces limitations to their style transfer effects. 
NNFM loss calculates the distance between each pixel in the rendered feature map and its nearest neighbor feature independently, without considering the relationships between pixels in the rendered feature map. This might cause the scene to independently learn local features rather than collaboratively learn global features.
Furthermore, acting as nearest neighbors, only a portion of the style features in the style feature map participate in the computation. These style features struggle to cover the entire style image, leading to further deprivation of style information.



Additionally, as an implicit representation, NeRF is difficult to edit controllably.
Recently, a novel 3D representation known as 3D Gaussian Splatting (3DGS)~\cite{kerbl3Dgaussians} has been introduced to address multiple tasks. We observe that using the explicit representation of Gaussian is beneficial for controllable editing~\cite{chen2023gaussianeditor}. In addition, 3DGS achieves higher rendering quality within a shorter training time and supports real-time rendering through a fast differentiable rasterizer. 

  
   
   

{In this paper, we introduce a novel stylization framework for 3D Gaussian Splatting~(ABC-GS) to enable different types of style transfer, including single-image, compositional, and semantic-aware. 
To this end, our ABC-GS consists of two stages: controllable matching and stylization.
Specifically, in the controllable matching stage, users can flexibly segment content images according to their needs and select style images and the style transfer type. 
Laying the foundation for obtaining high-quality style transfer results, we isolate the style features of different semantic labels and perform consistent color matching of the content image and scene Gaussians based on the mask-matching results.
In the stylization stage, we propose the Feature Alignment Style Transfer (FAST) loss to address the shortcomings of NNFM loss.
FAST Loss calculates the target features by aligning the rendered feature distribution with the style feature distribution in the image feature space. 
We also introduce depth loss and regularization terms to preserve the original geometric information while stylizing the scene's appearance. 
Comprehensive experimental results demonstrate that our ABC-GS approach offers enhanced controllability during style transfer.
The main innovations of our work can be summarized as follows:}


\begin{itemize}
\item {We introduce ABC-GS, a novel controllable 3D style transfer framework to enable multiple types style transfer with a designed controllable matching stage.}
\item {We propose a FAST loss to enable the stylization of 3D scenes to align with {the global style of the reference image faithfully}. }
\item {Experiments demonstrate that our method can achieve more controllable, high-quality stylization results in real-time rendering and ensure strict multi-view consistency.}
\end{itemize}

   
   
\section{Related Work}

\paragraph{Image Style Transfer.}
Gatys et al. \cite{gatys2016image} {first propose} the neural style transfer field by applying the style features of one image to another using the convolutional neural network (CNN). 
This pioneering work demonstrates that deep neural networks could capture style information from an image and successfully transfer it to another image. 
CNN-based style transfer techniques are primarily divided into optimization-based methods~\cite{chen2016fast, li2016combining} and feed-forward methods~\cite{huang2017arbitrary, huo2021manifold}.

\paragraph{3D Style Transfer.}
With the development of 2D style transfer and 3D representation, stylization is advanced towards the 3D world. 
Prior attempts at 3D style transfer are made on point cloud~\cite{huang2021learning} and mesh~\cite{hollein2022stylemesh}. Subsequently, various style transfer pipelines based on NeRF~\cite{liu2023stylerf, zhang2022arf, zhang2023ref}, demonstrate NeRF's superiority for 3D stylization tasks. {However, these methods lack fine-grained control over the resulting scenes. Recently, some 3D style transfer methods based on 3DGS~\cite{liu2023stylegaussian, zhang2024stylizedgs, kovacs2024g} have emerged. {These methods either use NNFM loss or pre-trained feed-forward models. Consequently, it is difficult to achieve high-quality stylization results that are faithful to the global style of the reference.} } Differently, we design the FAST loss that aligns features in the feature space for global stylization. Our work uses the image as a guide and aims to leverage the explicit characteristics of 3DGS for controllable stylization.
\section{Preliminary}
3D Gaussian Splatting~\cite{kerbl3Dgaussians} represents a 3D scene by a set of explicit 3D Gaussians $G=\{g_i\}$ where $g_i=(\mu_i, \Sigma_i, \sigma_i, c_i)$. Each Gaussian $g_i$ is parameterized by its mean $\mu_i$, covariance matrix $\Sigma_i$, opacity $\sigma_i$ and color $c_i$ which is represented in the coefficients of a spherical harmonic function. The covariance matrix can also be decomposed into rotation parameters $r_i$ and scale parameters $s_i$. 3D Gaussians can be effectively rendered by the fast differentiable rasterizer. Specifically, the color $C$ of a pixel is computed by blending the ordered Gaussians that overlap the pixel as
\begin{equation}
    C = \sum_{i \in {N}} T_i \alpha_i c_i, T_i=\prod_{j=1}^{i-1} (1 - \alpha_j),
\end{equation}
where $T_i$ is the transmittance and $\alpha_i$ is the alpha-compositing weight accumulated by $\Sigma_i$ and $\sigma_i$ for the $i$-th Gaussian. Please refer to \cite{kerbl3Dgaussians} for more details.
\section{Method}
\subsection{Overview}
{In the following section, we provide a detailed presentation of the proposed ABC-GS framework. The pipeline is shown in Fig.\ref{fig:pipeline}. Sec.\ref{CMS} introduces the econtrollable matching stage before scene stylization. Sec.\ref{ST} introduces our proposed novel 3D style transfer loss named feature alignment style transfer (FAST) loss. Additionally, we provide a detailed introduction of content and geometric loss, which better preserves the geometric content while implementing scene style transfer.}

\subsection{Controllable Matching Stage}
\label{CMS}
\paragraph{Mask Match}
{
We achieve matching between the content image, style image, and Gaussians within the scene through mask matching. Specifically, users can obtain the content mask $M_c$ using the SAM~\cite{kirillov2023segment}.} To ensure consistency in subsequent computational results, we utilize the explicit properties of 3D Gaussians to obtain semantic Gaussians. Similar to \cite{chen2023gaussianeditor}, we unproject the semantic labels of the content mask {$\{M_c^j\}_{j=1}^{z}$} onto the 3D Gaussians. For the $i$-th Gaussian $g_i$, $w_i^j$ represents the weight of its $j$-th semantic label, 
\begin{equation}
    w_i^j = \sum \alpha_i(p) \cdot T_i^j(p) \cdot M^j_c(p),
\end{equation}
where $p$ is the pixel of mask $M_c^j$, and $M_c^j$ corresponds to the $j$-th semantic label for content mask {$M_c$}. Next, by comparing the set threshold with the calculated weights, the semantic labels corresponding to the Gaussians can be determined. 
{Simultaneously, we create style mask $M_s$ for each style region. The style masks and content masks can be matched through semantic relationships or manually by the user. Once matched, the style masks also align with the Gaussians of the corresponding semantic labels in the content masks.} We define the $z$-th semantic matching group as $\Omega^z = \{M_s^z, M_c^z, G^z\}$, which will be applied in the computation of subsequent linear color transformation matrix and affinity matrix.

\begin{figure}[t]
  \centering
  
   \includegraphics[width=1\linewidth]{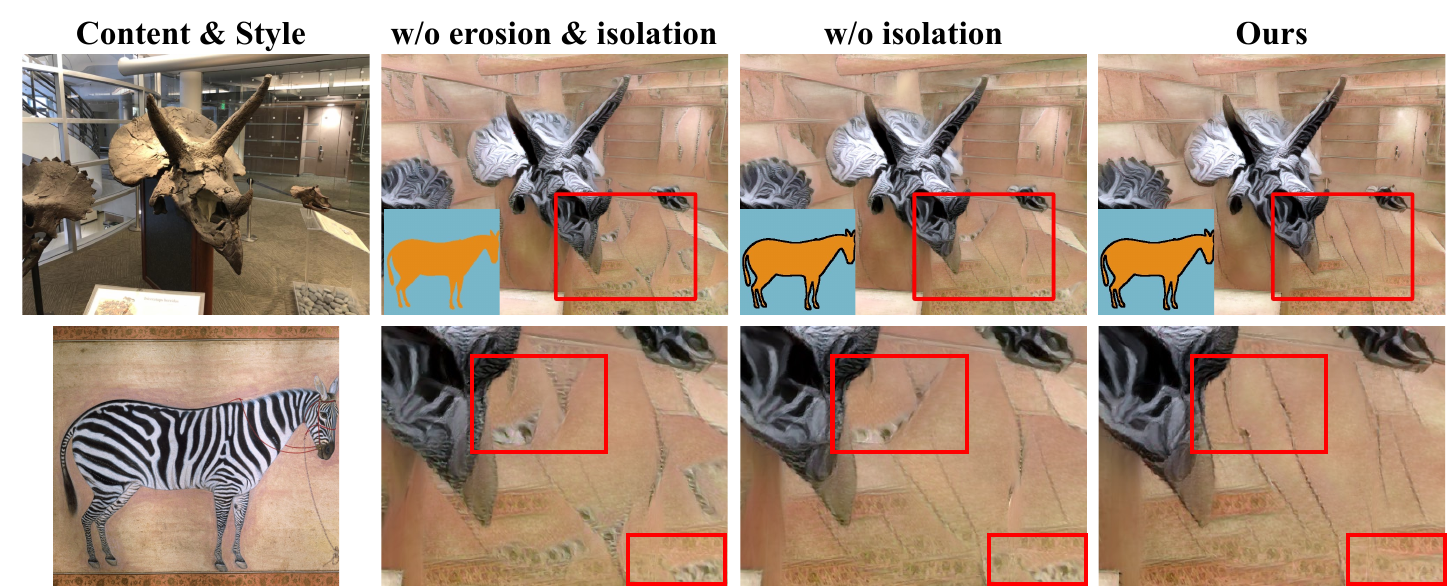}
   
   \caption{\textbf{Style Isolation.} {Only using either the style mask or the eroded style mask fails to prevent the leakage of the zebra texture style. Employing style isolation can effectively address this issue.}} 
   \vspace{-4mm}
   \label{fig:style_isolation}
\end{figure}

\paragraph{{Style Isolation}}
{For single-image and compositional style transfer, the style image mask covers the entire picture. However, semantic-aware style transfer requires extracting multiple semantic label masks within one style image. Previous methods~\cite{pang2023locally, zhang2024coarf} directly use the style image masks extracted by~\cite{kirillov2023segment}. We found that the boundaries between different semantic labels in the segmentation mask are not entirely accurate, which may lead to style leaking from one area to another. To address this, we perform erosion on the area of each semantic label in the style mask. However, as shown in Fig.\ref{fig:style_isolation}, the eroded masks still cannot achieve style isolation. Due to the use of the VGG-16 network~\cite{gatys2016image}, which has multiple convolution layers to extract style features, simply using feature map masks obtained by downsampling through pixel-level masks still contains style information from other style areas. To address this issue, we extract style features using the following method: for the style mask $M_s^z$ of the $z$-th semantic matching group $\Omega^z$, we isolate the pixels of this mask area into a new image and then complete the image through mirroring, translation, or color filling. The features $F_s^z$ from the mask area are used as the style features corresponding to the style mask $M_s^z$. This method effectively achieves style isolation.}

\paragraph{Color Match}
To enhance the stylization effects, it is desired that the colors of the stylized scene align with those of the style image. Inspired by \cite{zhang2022arf}, we perform a linear color transformation on the content images and Gaussians. Let $\{ p^{c}_i \}_{i=1}^{n}$ be the set of all pixel colors in the content images to be recolored, and let $\{ p^{s}_i \}_{i=1}^{m}$ be the set of all pixel colors of the style image. Specifically, we compute the linear color transformation weight and bias by matching the means and covariances of the content color set $\{p^{c}\}$ and the style color set $\{p^{s}\}$ as
\begin{equation}
\begin{aligned}
p^{ct} &= Ap^c + b, c^{ct} = Ac + b, \\
\text{s.t.} \quad E[p^{ct}] &= E[p^s], \text{Cov}[p^{ct}] = \text{Cov}[p^s],
\end{aligned}
\end{equation}
where $A \in \mathbb{R}^{{3} \times 3}$ and $b \in \mathbb{R}^{{3}}$ are the parameters required for linear color transformation, $c$ is the original Gaussian color, $p^{ct}$ and $c^{ct}$ respectively represent the pixel colors of the content image and the Gaussian colors after the linear color transformation. 
After the color transformation, the Gaussian-expressed scene and content images are not consistent. We further retrain the Gaussian parameters using the reconstruction loss described in \cite{kerbl3Dgaussians}. Specifically, the Gaussian parameters are optimized by calculating the $\mathcal{L}_1$ loss and $\mathcal{L}_{D-SSIM}$ loss between the rendered images and the content images as
\begin{equation}
    \mathcal{L}_{rec} = (1 - \lambda) \mathcal{L}_1 + \lambda \mathcal{L}_{D-SSIM}. \label{rec_loss}
\end{equation}

\begin{figure}[t]
  \centering
  
   \includegraphics[width=1\linewidth]{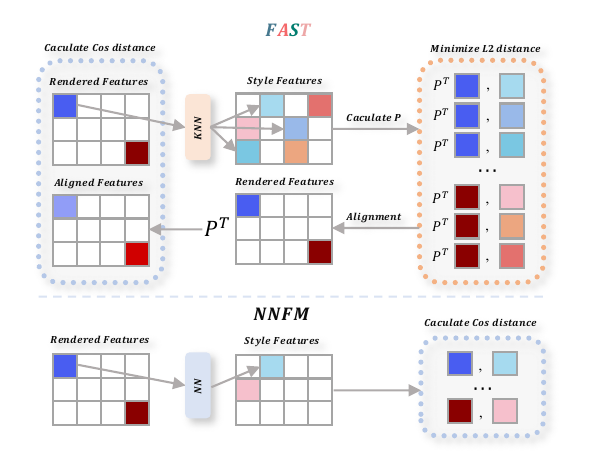}
   
   \caption{\textbf{FAST Loss and NNFM loss.} For the calculation of FAST loss, it first tallys all pairs of k-nearest neighbors between the rendered features and the style features, which are jointly used to compute the alignment matrix $P$. Style transfer is achieved by minimizing the cosine distance between the rendered features and the aligned features. The NNFM loss directly implements style transfer by minimizing the cosine distance between each rendered feature and its nearest neighbor in the style feature.} 
   
   \label{fig:fast}
   \vspace{-4mm}
\end{figure}

\subsection{Stylization Stage}
\label{ST}
\paragraph{Feature Alignment Style Transfer Loss}
The overall framework of the proposed FAST loss is given in Fig.\ref{fig:fast}. 
For the rendered image $I_r$, define their features extracted by the same feature extractors as $F_r$. Based on the semantic matching group $\Omega^z$ obtained through preprocessing, we extract the sets of feature vectors $F_r^z$ by $M_c^z$. To align feature distribution $F_r^z$ to $F_s^z$, an alignment matrix $P_z$ is computed to transform the rendered features.


We aim to prioritize the alignment of semantically similar features between the two feature distributions to achieve reasonable style transfer effects. We use the normalized similarity to build an affinity matrix, which aims to compensate for the inadequacy of semantic information in image features. Specifically, we denote the affinity matrix of the rendered features and the style features as $A^{z} \in \mathbb{R}^{N_r^z \times N_s^z}$, where $N_r^z$ and $N_s^z$ are the numbers of feature vectors in the rendered feature set and the style feature set of the z-th semantic matching group, respectively. Each element in $A^z$ is determined according to the following formula,
\begin{equation}
    A^{z}_{ij} = 
    \begin{cases} 
    1 & \text{if } v_{r}^{z,i} \in \mathcal{N}_k(v_{s}^{z,j}) \text{ or } v_{s}^{z,j} \in \mathcal{N}_k(v_{r}^{z,i}), \\
    0 & \text{otherwise},
    \end{cases}
    \label{equ:affinity matrix}
\end{equation}
where $v_{r}^{z,i}$ denotes the $i$-th feature from $F_r^z$ and $v_{s}^{z,j}$ is defined similarly. $\mathcal{N}_k(v)$ is a set of $v$'s $k$-nearest neighbors in the other feature set without $v$. The neighbors are found by the normalized similarity. {We employ all neighborhood relationships to compute the alignment matrix $P_z$, which is used to derive the aligned features $F_{rs}^z$. This calculation method takes into account the pixels of rendered feature maps, which is conducive to learning global style.}
We achieve feature alignment by bringing similar features closer together within the two feature distributions in the feature subspace. The objective function can be articulated as follows,
\begin{equation}
    P_z = \mathop{\arg\min}\limits_{P_z} \frac{1}{N_{pair}^z} \sum_{i=1}^{N_r^z} \sum_{j=1}^{N_s^z} A_{ij}^{z} \left\| {P_z^\mathrm{T}}v_r^{z,i} - v_s^{z,j} \right\|^2_2,
    \label{equ:align}
\end{equation}
where $N_{pair}^z$ is the number of pairs of the nearest neighbors for $z$-th senmatic label, $N_r^z$ and $N_s^z$ are the numbers of feature vectors in $F_r^z$ and $F_s^z$, respectively. Through Eq.\ref{equ:align}, we can calculate the projection matrix $P_z$. Please refer to the supplementary materials for the mathematical derivation. We can obtain the feature $F_{rs}^z$, aligned to the style feature distribution, from the formula $F_{rs}^z=P_z^\mathrm{T}F_r$. By applying the aforementioned transformation to all rendered feature vectors according to their semantic matching groups, we can obtain the aligned feature map $F_{rs}$. Loss function can be written as
\begin{equation}
    \mathcal{L}_{FAST} (F_{r}, F_{rs}) = \frac{F_{r} \cdot F_{rs}}{\|F_{r}\| \|F_{rs}\|}.
\end{equation}

\paragraph{Content and Geometric Protection Loss}
Focusing solely on stylization will make it difficult to recognize the original content in the scene. To address this issue, we add a content preservation loss $L_{con}$ as
\begin{equation}
    \mathcal{L}_{con} = \frac{1}{N_F} \|F_{c} - F_{r}\|_2^2,
\end{equation}
where $N_F$ is the number of pixels in the rendered feature map. Additionally, we compute the total variation loss $L_{tv}$ of the rendered image to reduce high-frequency noise in the image while preserving its overall structure. 

Furthermore, a depth loss is proposed to preserve the geometric information contained within the Gaussians in the scene. Similar to computing the color, the depth value $D$ of a pixel is computed by blending $N$ ordered Gaussians overlapping the pixel,  
\begin{equation}
    D = \sum_{i \in {N}} T_i \alpha_i d_i,
\end{equation}
where $d_i$ is the depth for 
$i$-th Gaussian. {We use the original Gaussians to render the initial depth map $D_{init}$.} Let $D_r$ denotes the rendered depth map, the depth loss is defined as
\begin{equation}
    \mathcal{L}_{dep} = \frac{1}{N_D} \| D_{init} - D_{r} \|_2^2,
\end{equation}
where $N_D$ is the number of depth map pixels. 

In addition, we incorporate the scale and opacity regularization terms {calculated with the original Gaussian parameters} to alleviate the blurred or needle-like effect caused by Gaussian overfitting, 
\begin{equation}
    \mathcal{L}_{sca} = \|\Delta s\|_2, \quad 
    \mathcal{L}_{opa} = \|\Delta \alpha\|_2.
\end{equation}
\begin{figure*}[t]
   \centering
  
   \includegraphics[width=1\linewidth]{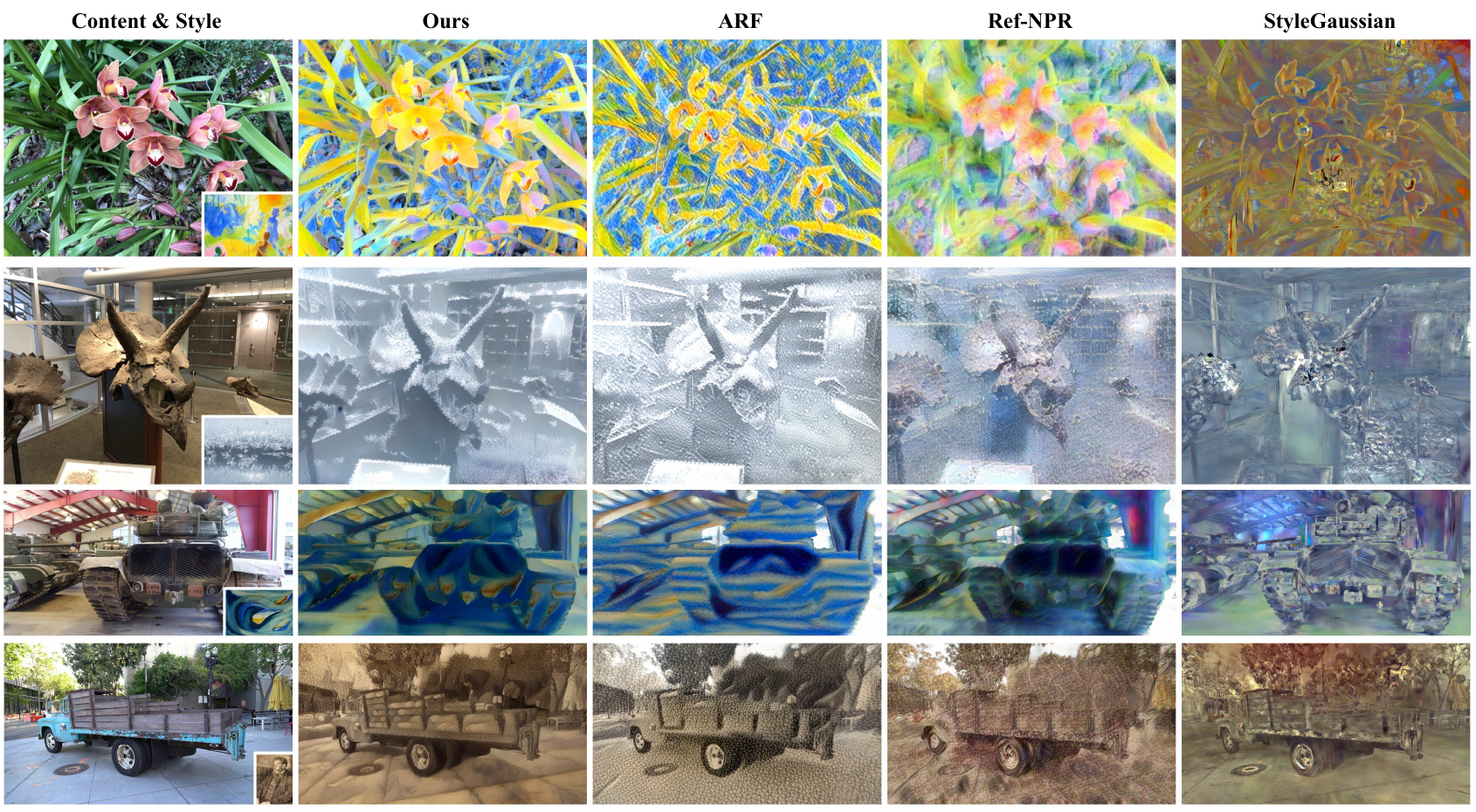}
   
   \caption{\textbf{Qualitative comparisons with the baseline methods on LLFF and T\&T datasets.} Compared to other methods, our method better preserves the geometric information of the scene while achieving style transfer.} 
   \vspace{-3mm}
   \label{fig:qual_llff}
\end{figure*}

  
   
   

  
   
   

\section{Experiments}
We select two real-world scene datasets, LLFF and T\&T, for our experiments. {In addition, we use the WikiArt dataset~\cite{saleh2015large} and ARF style dataset as the style image dataset. For single-image style transfer, we compare our method with the state-of-the-art 3D style transfer methods, including ARF~\cite{zhang2022arf}, Ref-NPR~\cite{zhang2023ref} and StyleGaussian~\cite{liu2023stylegaussian}. Since Ref-NPR is a reference-based method, we utilize AdaIN~\cite{huang2017arbitrary} to obtain its stylized 2D views before styling the scene. 
Pretrained VGG-16 model~\cite{simonyan2014very} layers in conv3 block are used to extract features. We set k to 5 in Eq.\ref{equ:affinity matrix} to achieve style transfer while preventing the results from being overly smooth. For the loss during the stylization stage, we set the weight $\lambda^{*}=\{2, 0.005, 0.02, 0.01, 1, 1\}$ for the loss functions $\{L_{FAST}, L_c, L_{tv}, L_{dep}, L_{sca}, L_{opa}\}$ and disable the densification strategy. All our experiments are conducted on a single NVIDIA RTX 4090 GPU.

\subsection{Qualitative Evaluation}
We conduct a qualitative comparison of single-image style transfer between our method and the baseline, as shown in Fig.\ref{fig:qual_llff}. StyleGaussian~\cite{liu2023stylegaussian}, due to using AdaIN~\cite{huang2017arbitrary} for arbitrary style transfer, struggles to achieve high-quality stylization based on a specified style image. 
ARF~\cite{zhang2022arf} and Ref-NPR~\cite{zhang2023ref} show positive progress in learning style features. However, since the appearance and geometry of the neural radiance field-based methods are difficult to disentangle, the content information and geometric information of the original scene are severely lost during the process of style transfer. In contrast, our method can better preserve the content information of the original scene, while the style transfer effect is faithful to {the global style of the reference image}. Fig.\ref{fig:others} presents the stylization results of the other style transfer type, which prove the controllability of ABC-GS.

\begin{figure}[t]
  \centering
  
   \includegraphics[width=1\linewidth]{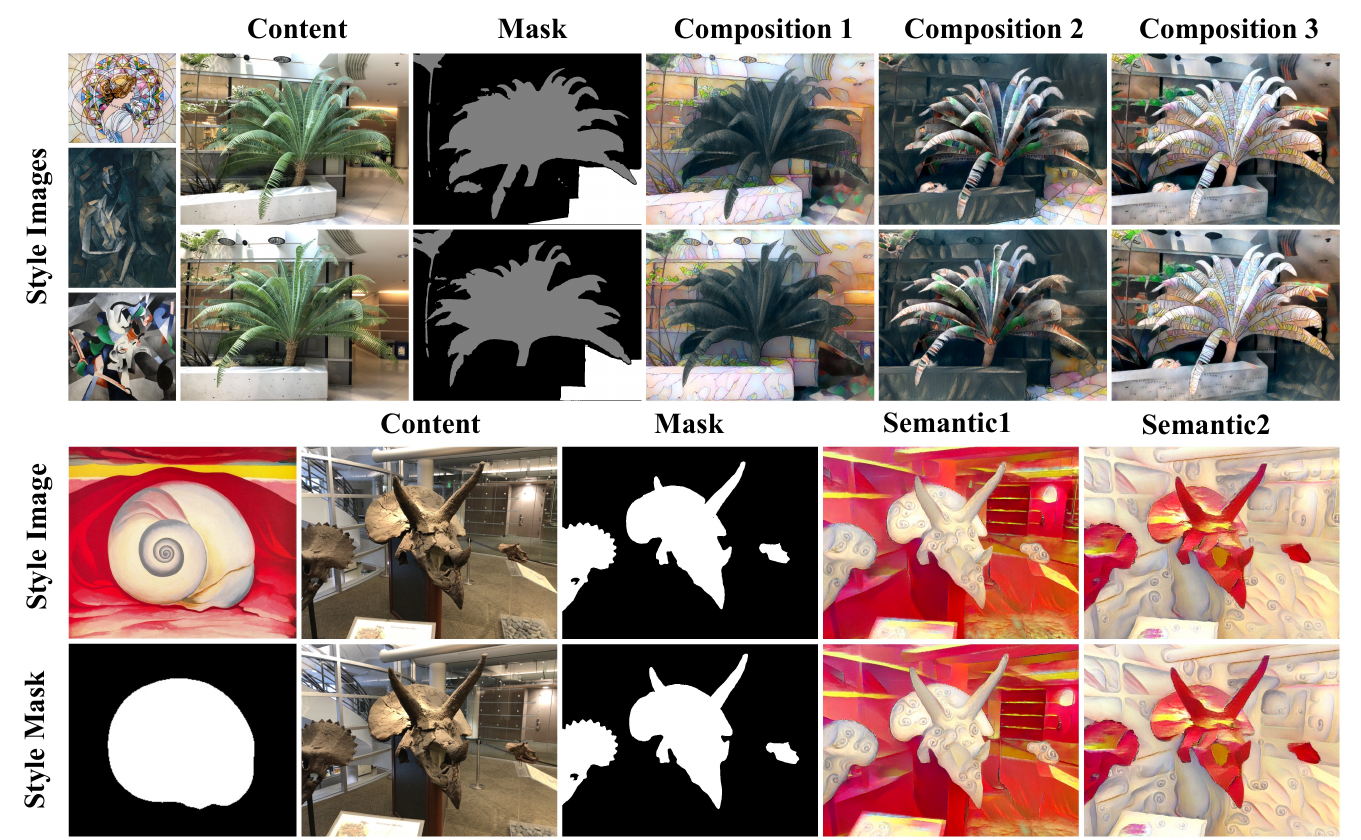}
   
   \caption{\textbf{Stylization result of compositional and semantic-aware style transfer.} Our approach enables controllable style transfer.} 
   \vspace{-3mm}
   \label{fig:others}
\end{figure}

\begin{table}[t]
\centering
\caption{\textbf{Comparison of various methods on style transfer performance and efficiency.} Among them, StyleGaussian utilizes a pre-trained model for real-time style transfer, so its training time is not accounted for.}
\resizebox{0.48\textwidth}{!}{
\begin{tabular}{lcccc}
\toprule
Metrics     & ArtFID ($\downarrow$) & SSIM ($\uparrow$) & Train Time($\downarrow$) &  FPS ($\uparrow$) \\
\midrule
StyleGaussian  & 47.45 & 0.33  & - & 9.7 \\
ARF            & 47.36 & 0.28  & 1.83 & 8.4 \\
Ref-NPR        & 45.85 & 0.30  & 2.28 & 7.1 \\
Ours           & \textbf{42.80} & \textbf{0.54}  & \textbf{1.60} & \textbf{153} \\
\bottomrule
\end{tabular}
}
 \vspace{-5mm}
\label{tab:artfid}
\end{table}

\begin{table}[t]
\centering
\caption{\textbf{Comparative assessment of methods for style transfer consistency.} From the results, it can be seen that our method achieves strict multi-view consistency.}
\resizebox{0.48\textwidth}{!}{
\begin{tabular}{l c c c c c}
\toprule
\textbf{Methods}  & \multicolumn{2}{c}{\textbf{Short-term Consistency}} & \multicolumn{2}{c}{\textbf{Long-term Consistency}} \\
                 & LPIPS($\downarrow$) & RMSE($\downarrow$) & LPIPS($\downarrow$) & RMSE($\downarrow$) \\
\midrule
StyleGaussian          & 0.064 & 0.058 & 0.109 & 0.089 \\
ARF                   & 0.111 & 0.082 & 0.173 & 0.120 \\
Ref-NPR                 & 0.099 & 0.074 & 0.148 & 0.104 \\
Ours                  & \textbf{0.057} & \textbf{0.048} & \textbf{0.093} & \textbf{0.077} \\
\bottomrule
\end{tabular}
}
\vspace{-4mm}
\label{tab:consistency}
\end{table}

\subsection{Quantitative Evaluation}

\paragraph{Performance and Efficiency}
For 3D scene stylization, the speed and quality of style transfer are crucial. We perform a quantitative comparison with the state-of-the-art methods on the LLFF dataset. Specifically, we use ArtFID to evaluate the stylization quality and use SSIM between the content images and the stylized images to judge the degree to which the style image retains the original scene information. The results are shown in Table.\ref{tab:artfid}. It can be seen that our method is capable of achieving high-quality style transfer effects while preserving the geometric information of the scene content. Additionally, our framework is based on 3DGS, enabling rapid training and real-time rendering.

\paragraph{Multi-View Consistency}
For multi-view consistency, we use optical flow to warp one view to another, and then calculate the RMSE score and LPIPS score to measure the multi-view consistency of the stylized scene. As shown in Table.\ref{tab:consistency}, ABC-GS outperforms previous methods. {Due to excessive loss of geometric information during the style transfer process, NeRF-based method exhibit poor short-term and long-term consistency.} In contrast, our method preserves the geometric information through depth loss and regularization terms, achieving better consistency.

\paragraph{User Study}
{We randomly select 15 groups of stylized scenes from the LLFF and T\&T datasets for comparison. For each comparison group, we set the evaluation grades from three dimensions: Visual Effect, Content Preservation Quality, {and Style Transfer Quality}. We recruit 32 users to evaluate each dimension of stylized scenes. The results are shown in Fig.\ref{fig:User Study}. Compared to the baseline, voters prefer our method from various perspectives. }


\begin{figure}[t]
  \centering
  
   \includegraphics[width=1\linewidth]{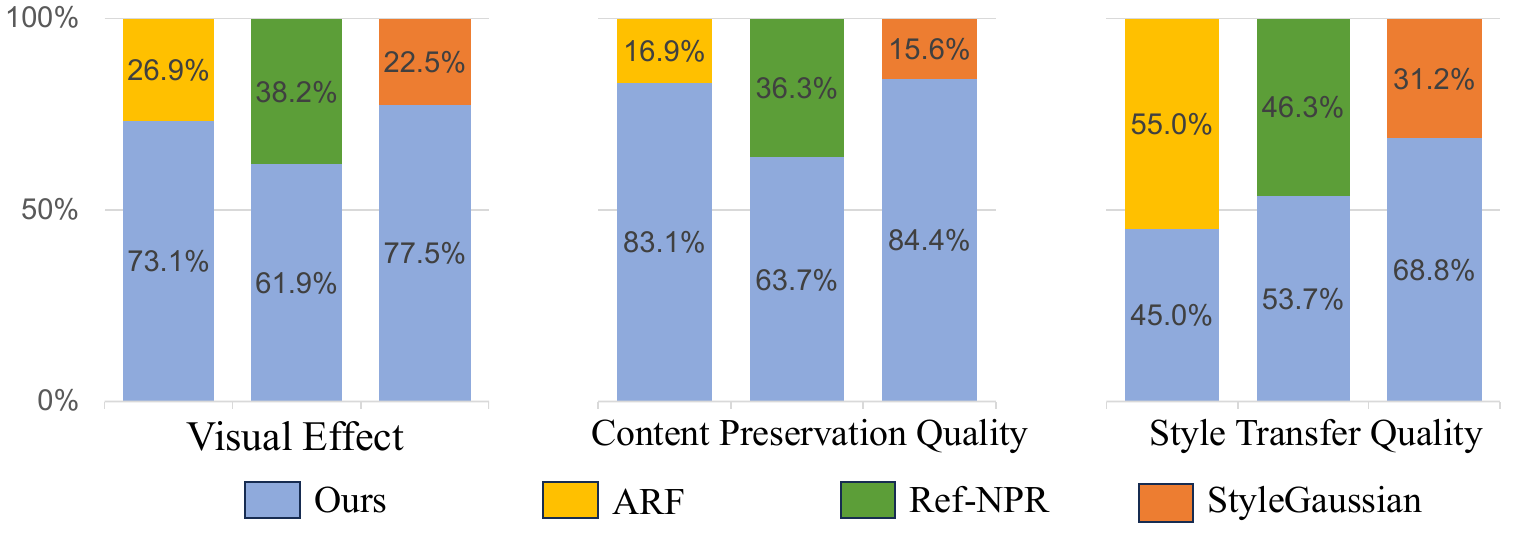}
   
   \caption{\textbf{User Study.} We conduct multi-faceted comparisons against the baseline, and the results show that our style transfer results are more preferred by users.} 
   \vspace{-3mm}
   \label{fig:User Study}
\end{figure}

\begin{figure}[t]
  \centering
  
   \includegraphics[width=1\linewidth]{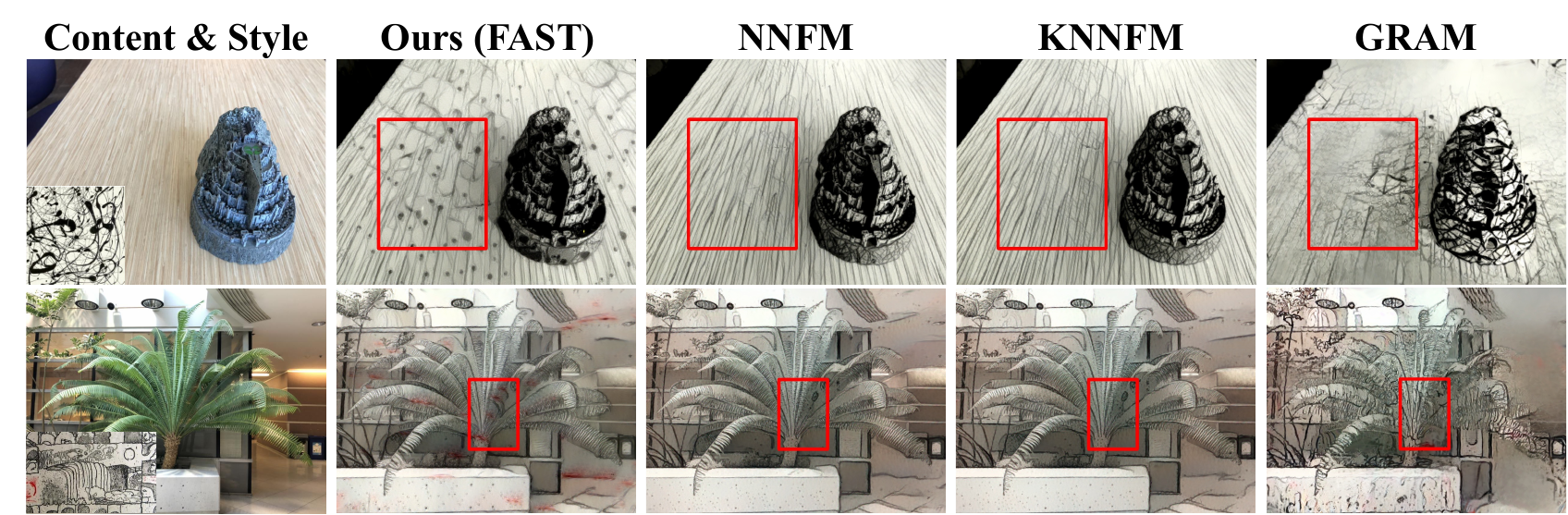}
   
   \caption{\textbf{Ablation Study for loss function.}  Our method transfers global style, thereby preventing the loss of style elements such as texture and color.}
   \vspace{-4mm}
   \label{fig:ablation loss}
\end{figure}

\subsection{Ablation Study}
{We compare our FAST loss to the NNFM loss~\cite{zhang2022arf}, KNNFM loss and Gram loss~\cite{gatys2016image}. In the case of KNNFM loss, the nearest neighbor approach is modified to K-nearest neighbors, and an average is taken for the calculations. K is same as k used previously in FAST loss. { As shown in }Fig.\ref{fig:ablation loss}, our FAST loss significantly transfers the global style features of the image. However, the results of KNNFM and NNFM are roughly the same, indicating that merely using K-nearest neighbors without considering the relationships between pixels in the rendered feature image does not yield better results. Gram loss leads to results with more artifacts.}

\section{Conclusion}
\label{Conclusion}
{In this paper, we proposed a controllable 3D style transfer framework named ABC-GS, which enabled single-image, compositional, and semantic-aware style transfer. We exploited the explicit properties of 3D Gaussian to design a controllable matching stage, which can achieve matching between scene content and style regions for fine-grained style transfer. In addition, our proposed FAST loss achieved a style transfer effect that is more faithful to the global style of the style image. To preserve geometric information while performing style transfer, we introduced depth loss and regularization terms in the stylization stage. Extensive experiments demonstrated the effectiveness of our method. }

\bibliographystyle{IEEEbib}
\bibliography{icme2025references}

\begin{thebibliography}{10}

\bibitem{mildenhall2020nerf}
Ben Mildenhall, Pratul~P. Srinivasan, Matthew Tancik, Jonathan~T. Barron, Ravi Ramamoorthi, and Ren Ng,
\newblock ``Nerf: Representing scenes as neural radiance fields for view synthesis,''
\newblock in {\em ECCV}, 2020.

\bibitem{huang2022stylizednerf}
Yi-Hua Huang, Yue He, Yu-Jie Yuan, Yu-Kun Lai, and Lin Gao,
\newblock ``Stylizednerf: consistent 3d scene stylization as stylized nerf via 2d-3d mutual learning,''
\newblock in {\em Proceedings of the IEEE/CVF Conference on Computer Vision and Pattern Recognition}, 2022, pp. 18342--18352.

\bibitem{liu2023stylerf}
Kunhao Liu, Fangneng Zhan, Yiwen Chen, Jiahui Zhang, Yingchen Yu, Abdulmotaleb El~Saddik, Shijian Lu, and Eric~P Xing,
\newblock ``Stylerf: Zero-shot 3d style transfer of neural radiance fields,''
\newblock in {\em Proceedings of the IEEE/CVF Conference on Computer Vision and Pattern Recognition}, 2023, pp. 8338--8348.

\bibitem{zhang2022arf}
Kai Zhang, Nick Kolkin, Sai Bi, Fujun Luan, Zexiang Xu, Eli Shechtman, and Noah Snavely,
\newblock ``Arf: Artistic radiance fields,''
\newblock in {\em European Conference on Computer Vision}. Springer, 2022, pp. 717--733.

\bibitem{zhang2023ref}
Yuechen Zhang, Zexin He, Jinbo Xing, Xufeng Yao, and Jiaya Jia,
\newblock ``Ref-npr: Reference-based non-photorealistic radiance fields for controllable scene stylization,''
\newblock in {\em Proceedings of the IEEE/CVF Conference on Computer Vision and Pattern Recognition}, 2023, pp. 4242--4251.

\bibitem{zhang2024coarf}
Deheng Zhang, Clara Fernandez-Labrador, and Christopher Schroers,
\newblock ``Coarf: Controllable 3d artistic style transfer for radiance fields,''
\newblock {\em arXiv preprint arXiv:2404.14967}, 2024.

\bibitem{kerbl3Dgaussians}
Bernhard Kerbl, Georgios Kopanas, Thomas Leimk{\"u}hler, and George Drettakis,
\newblock ``3d gaussian splatting for real-time radiance field rendering,''
\newblock {\em ACM Transactions on Graphics}, vol. 42, no. 4, July 2023.

\bibitem{chen2023gaussianeditor}
Yiwen Chen, Zilong Chen, Chi Zhang, Feng Wang, Xiaofeng Yang, Yikai Wang, Zhongang Cai, Lei Yang, Huaping Liu, and Guosheng Lin,
\newblock ``Gaussianeditor: Swift and controllable 3d editing with gaussian splatting,''
\newblock {\em arXiv preprint arXiv:2311.14521}, 2023.

\bibitem{gatys2016image}
Leon~A Gatys, Alexander~S Ecker, and Matthias Bethge,
\newblock ``Image style transfer using convolutional neural networks,''
\newblock in {\em Proceedings of the IEEE conference on computer vision and pattern recognition}, 2016, pp. 2414--2423.

\bibitem{chen2016fast}
Tian~Qi Chen and Mark Schmidt,
\newblock ``Fast patch-based style transfer of arbitrary style,''
\newblock {\em arXiv preprint arXiv:1612.04337}, 2016.

\bibitem{li2016combining}
Chuan Li and Michael Wand,
\newblock ``Combining markov random fields and convolutional neural networks for image synthesis,''
\newblock in {\em Proceedings of the IEEE conference on computer vision and pattern recognition}, 2016, pp. 2479--2486.

\bibitem{huang2017arbitrary}
Xun Huang and Serge Belongie,
\newblock ``Arbitrary style transfer in real-time with adaptive instance normalization,''
\newblock in {\em Proceedings of the IEEE international conference on computer vision}, 2017, pp. 1501--1510.

\bibitem{huo2021manifold}
Jing Huo, Shiyin Jin, Wenbin Li, Jing Wu, Yu-Kun Lai, Yinghuan Shi, and Yang Gao,
\newblock ``Manifold alignment for semantically aligned style transfer,''
\newblock in {\em Proceedings of the IEEE/CVF International Conference on Computer Vision}, 2021, pp. 14861--14869.

\bibitem{huang2021learning}
Hsin-Ping Huang, Hung-Yu Tseng, Saurabh Saini, Maneesh Singh, and Ming-Hsuan Yang,
\newblock ``Learning to stylize novel views,''
\newblock in {\em Proceedings of the IEEE/CVF International Conference on Computer Vision}, 2021, pp. 13869--13878.

\bibitem{hollein2022stylemesh}
Lukas H{\"o}llein, Justin Johnson, and Matthias Nie{\ss}ner,
\newblock ``Stylemesh: Style transfer for indoor 3d scene reconstructions,''
\newblock in {\em Proceedings of the IEEE/CVF Conference on Computer Vision and Pattern Recognition}, 2022, pp. 6198--6208.

\bibitem{liu2023stylegaussian}
Kunhao Liu, Fangneng Zhan, Muyu Xu, Christian Theobalt, Ling Shao, and Shijian Lu,
\newblock ``Stylegaussian: Instant 3d style transfer with gaussian splatting,''
\newblock {\em arXiv preprint arXiv:2403.07807}, 2024.

\bibitem{zhang2024stylizedgs}
Dingxi Zhang, Zhuoxun Chen, Yu-Jie Yuan, Fang-Lue Zhang, Zhenliang He, Shiguang Shan, and Lin Gao,
\newblock ``Stylizedgs: Controllable stylization for 3d gaussian splatting,''
\newblock {\em arXiv preprint arXiv:2404.05220}, 2024.

\bibitem{kovacs2024g}
{\'A}ron~Samuel Kov{\'a}cs, Pedro Hermosilla, and Renata~G Raidou,
\newblock ``g-style: Stylized gaussian splatting,''
\newblock in {\em Computer Graphics Forum}. Wiley Online Library, 2024, p. e15259.

\bibitem{kirillov2023segment}
Alexander Kirillov, Eric Mintun, Nikhila Ravi, Hanzi Mao, Chloe Rolland, Laura Gustafson, Tete Xiao, Spencer Whitehead, Alexander~C Berg, Wan-Yen Lo, et~al.,
\newblock ``Segment anything,''
\newblock in {\em Proceedings of the IEEE/CVF International Conference on Computer Vision}, 2023.

\bibitem{pang2023locally}
Hong-Wing Pang, Binh-Son Hua, and Sai-Kit Yeung,
\newblock ``Locally stylized neural radiance fields,''
\newblock in {\em 2023 IEEE/CVF International Conference on Computer Vision (ICCV)}. IEEE Computer Society, 2023, pp. 307--316.

\bibitem{saleh2015large}
Babak Saleh and Ahmed Elgammal,
\newblock ``Large-scale classification of fine-art paintings: Learning the right metric on the right feature,''
\newblock {\em arXiv preprint arXiv:1505.00855}, 2015.

\bibitem{simonyan2014very}
Karen Simonyan and Andrew Zisserman,
\newblock ``Very deep convolutional networks for large-scale image recognition,''
\newblock {\em arXiv preprint arXiv:1409.1556}, 2014.

\end{thebibliography}

\appendix
{The appendix provides additional details and in-depth analyses of ABC-GS. The code is released at \href{https://vpx-ecnu.github.io/ABC-GS-website}{https://vpx-ecnu.github.io/ABC-GS-website}. The structure of the appendix is as follows:}
{
\begin{itemize}
    \item Derivation of the color transformation.
    \item Derivation of the feature alignment.
    \item More Implementation Details.
    \item A more comprehensive ablation study. 
    \item Additional qualitative results.
    \item Limitations.
\end{itemize}
}

\subsection{{Derivation of the Color Transformation}}

{
First, we perform eigen-decompositions on the covariance matrices $\text{Cov}[p^c]$ and $\text{Cov}[p^s]$, }

\begin{equation}
\text{Cov}[{p^c}] = U_c \Lambda_c U_c^T,
\end{equation}

\begin{equation}
\text{Cov}[{p^s}] = U_s \Lambda_s U_s^T,
\end{equation}
where $U_c$ and $U_s$ are orthogonal matrices containing the eigenvectors, $\Lambda_c$ and $\Lambda_s$ are diagonal matrices containing the eigenvalues of the covariance matrices $\text{Cov}[{p^c}]$ and $\text{Cov}[{p^s}]$, respectively. 

Let $A$ denote the weights, and $b$ denote the bias, the solutions are as follows:
\begin{equation}
A = U_s \Lambda_s^{\frac{1}{2}} U_s^T U_c \Lambda_c^{-\frac{1}{2}} U_c^T,
\end{equation}

\begin{equation}
b = {E}[p^s] - A \cdot {E}[p^c].
\end{equation}


First, we prove that ${E}[p^{ct}] = {E}[p^s]$ holds true:

\begin{align}
    {E}[p^{ct}] &= {E}[Ap^c + b]\nonumber  \\
                 &= {E}[Ap^c] + {E}[b]\nonumber  \\
                 &= A \cdot {E}[p^c] + b\nonumber  \\
                 &= A \cdot {E}[p^c] + ({E}[p^s] - A \cdot {E}[p^c])\nonumber  \\
                 &= A \cdot {E}[p^c] + {E}[p^s] - A \cdot {E}[p^c]\nonumber  \\
                 &= {E}[p^s].
\end{align}

Next, we present the proof for $\text{Cov}[p^{ct}] = \text{Cov}[p^s]$:

\begin{align}
    \text{Cov}[p^{ct}] &= \text{Cov}[Ap^c + b] \nonumber \\  
                       &= \text{Cov}[Ap^c] \quad \text{(since $b$ is constant)} \nonumber  \\
                       &= A \text{Cov}[p^c] A^T \nonumber \\
                       &= (U_s \Lambda_s^{\frac{1}{2}} U_s^T U_c \Lambda_c^{-\frac{1}{2}} U_c^T) (U_c \Lambda_c U_c^T) \nonumber  \\
                       &\quad \; (U_c \Lambda_c^{-\frac{1}{2}} U_c^T U_s \Lambda_s^{\frac{1}{2}} U_s^T)^T \nonumber \\
                       &= U_s \Lambda_s^{\frac{1}{2}} U_s^T (U_c \Lambda_c^{-\frac{1}{2}} \Lambda_c \Lambda_c^{-\frac{1}{2}} U_c^T) U_s \Lambda_s^{\frac{1}{2}} U_s^T \nonumber \\
                       &= U_s \Lambda_s^{\frac{1}{2}} U_s^T (U_c U_c^T) U_s \Lambda_s^{\frac{1}{2}} U_s^T \nonumber \\
                       &= U_s \Lambda_s^{\frac{1}{2}} U_s^T U_s \Lambda_s^{\frac{1}{2}} U_s^T\nonumber  \\
                       &= U_s \Lambda_s U_s^T \nonumber  \\
                       &= \text{Cov}[p^s].
\end{align}

\subsection{{Derivation of the Feature Alignment}}


To solve the optimization problem of feature alignment, let
\begin{align}
    J(P_z) &= \frac{1}{N_{pair}^z} \sum_{i=1}^{N_r^z} \sum_{j=1}^{N_s^z} A_{ij}^{z} \left\| {P_z^\mathrm{T}}v_r^{z,i} - v_s^{z,j}\right\|^2_2, \nonumber\\
            &= \frac{1}{N_{pair}^z} \sum_{i=1}^{N_r^z} \sum_{j=1}^{N_s^z} A_{ij}^{z} \left( P_z^T v_r^{z,i} - v_s^{z,j} \right)^T \nonumber\\
            &\quad \left( P_z^T v_r^{z,i} - v_s^{z,j} \right) \nonumber\\
            &= \frac{1}{N_{\text{pair}}^z} \sum_{i=1}^{N_r^z} \sum_{j=1}^{N_s^z} A_{ij}^z \left[ (v_r^{z,i})^T P_z P_z^T v_r^{z,i} \nonumber\right.  \nonumber\\
            &\quad - 2 \left. (v_r^{z,i})^T P_z v_s^{z,j} + (v_s^{z,j})^T v_s^{z,j} \right].
\end{align}

To simplify the expression, we use the trace operator. The summation in matrix form is represented as follows:

\begin{align}
J(P_z) &= \text{tr}\left(P_z^T F_r^z D_r^z (F_r^z)^T P_z\right) + \text{tr}\left(F_s^z D_s^z (F_s^z)^T\right) \nonumber\\
&\quad - 2 \text{tr}\left(P_z^T F_r^z U_z (F_s^z)^T\right),
\end{align}
where $U_{z}=\frac{1}{N_{pair}^z} A^{z}$, $D_r^z \in \mathbb{R} ^ {{N_r^z} \times {N_r^z}}$ is a diagonal matrix, with $D_r^z(i,i) = \frac{1}{N_{pair}^z}\Sigma_{j=1}^{N_s^z} A_{ij}^z$. $D_s^z \in \mathbb{R} ^ {{N_s^z} \times {N_s^z}}$ is also a diagonal matrix, and $D_s^z(j,j) = \frac{1}{N_{pair}^z}\Sigma_{i=1}^{N_r^z} A_{ij}^z$.

 Since the style images do not change during the training process, the value of the second item is fixed. We take the derivative of $J(P_z)$ with respect to $P_z$:
\begin{equation}
    \frac{\partial J}{\partial P_z} = 2(F_r^z D_r^z (F_r^z)^T P_z - F_r^z U_z (F_s^z)^T).
\end{equation}

The objective is to find $P_z$ that minimizes $J(P_z)$. Therefore, we set the derivative to 0 to solve for the optimal solution,
\begin{equation}
    P_z = (F_r^z D_r^z (F_r^z)^T)^{-1} (F_r^z U_{z} (F_s^z)^T).
\end{equation}

\subsection{Implementation Details}
To prevent floating Gaussian coloring, we use outlier removal to filter out floating Gaussians when loading the original Gaussians and before starting the stylization stage. Additionally, during the training process of optimizing Gaussian colors in the controllable matching stage, we continuously filter out Gaussians that have low opacity or large scale. Non-photorealistic rendering generally does not require consideration of lighting information. To simplify the color transformation during the color matching stage, we default to training using zero-degree spherical harmonics coefficients. The scene already possesses a sufficient number of Gaussians. To prevent an excessive number of Gaussians, we disable the densification strategy during the stylization stage.




\subsection{{Ablation study}}

\paragraph{{Ablation Study for Color Match}}

{The Color match module is designed to align the colors of the stylized scene and the style image to enhance the stylization effect. As shown in Fig.\ref{fig:ablation_Ct}, the second column demonstrates the effects of removing this module. The significant color differences between these scenes and the style images reduce the visual impact of the stylization. }\

\begin{figure}[t]
  \centering

   \includegraphics[width=1\linewidth]{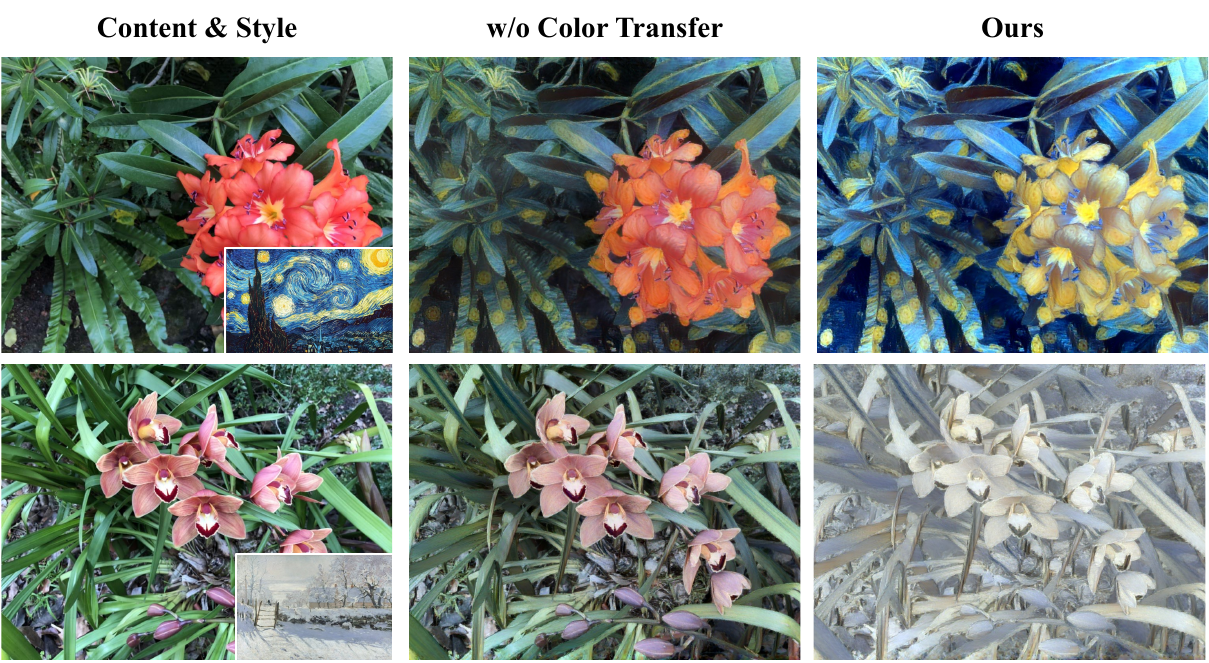}
   
   \caption{\textbf{{Ablation Study for Color Match.}} {We compare the qualitative results before and after removing the color match module.}} 
   \label{fig:ablation_Ct}
\end{figure}

\begin{figure}[t]
  \centering

   \includegraphics[width=1\linewidth]{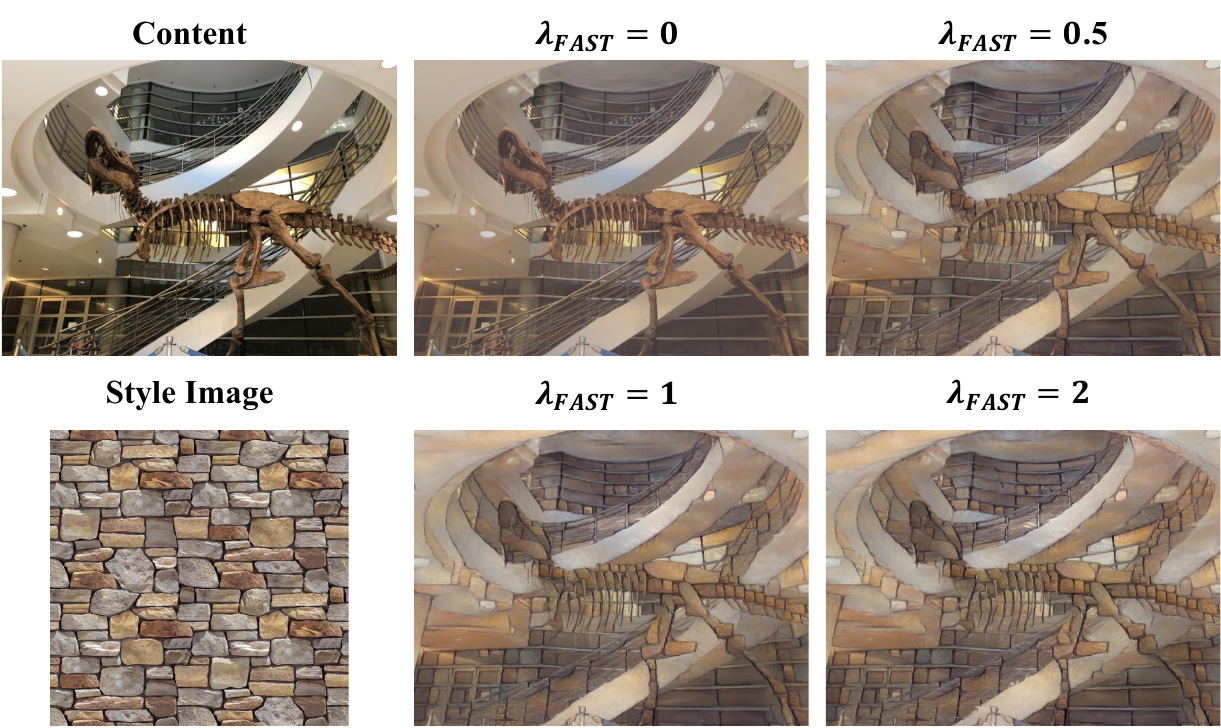}
   
   \caption{\textbf{{Ablation Studies on different $\lambda_{FAST}$.}} {$\lambda_{FAST}$ values of 0, 0.5, 1, and 2.}} 
   \label{fig:ablation_fast}
\end{figure}

\paragraph{Effectiveness of Parameter $\lambda_{FAST}$} 
In the Stylization Stage, $\lambda_{FAST}$ plays an important role in controlling the intensity of the style transfer. As shown in Fig.\ref{fig:ablation_fast}, evaluations are conducted to observe the stylization effects when $\lambda_{FAST}$ was set to 0, 0.5, 1, and 2. The smaller the coefficient, the weaker the degree of stylization. When the coefficient is 0, only the colors match with the style image. Users can adjust the parameter $\lambda_{FAST}$ according to their preferences to control the intensity of stylization.

\begin{figure}[t]
  \centering

   \includegraphics[width=1\linewidth]{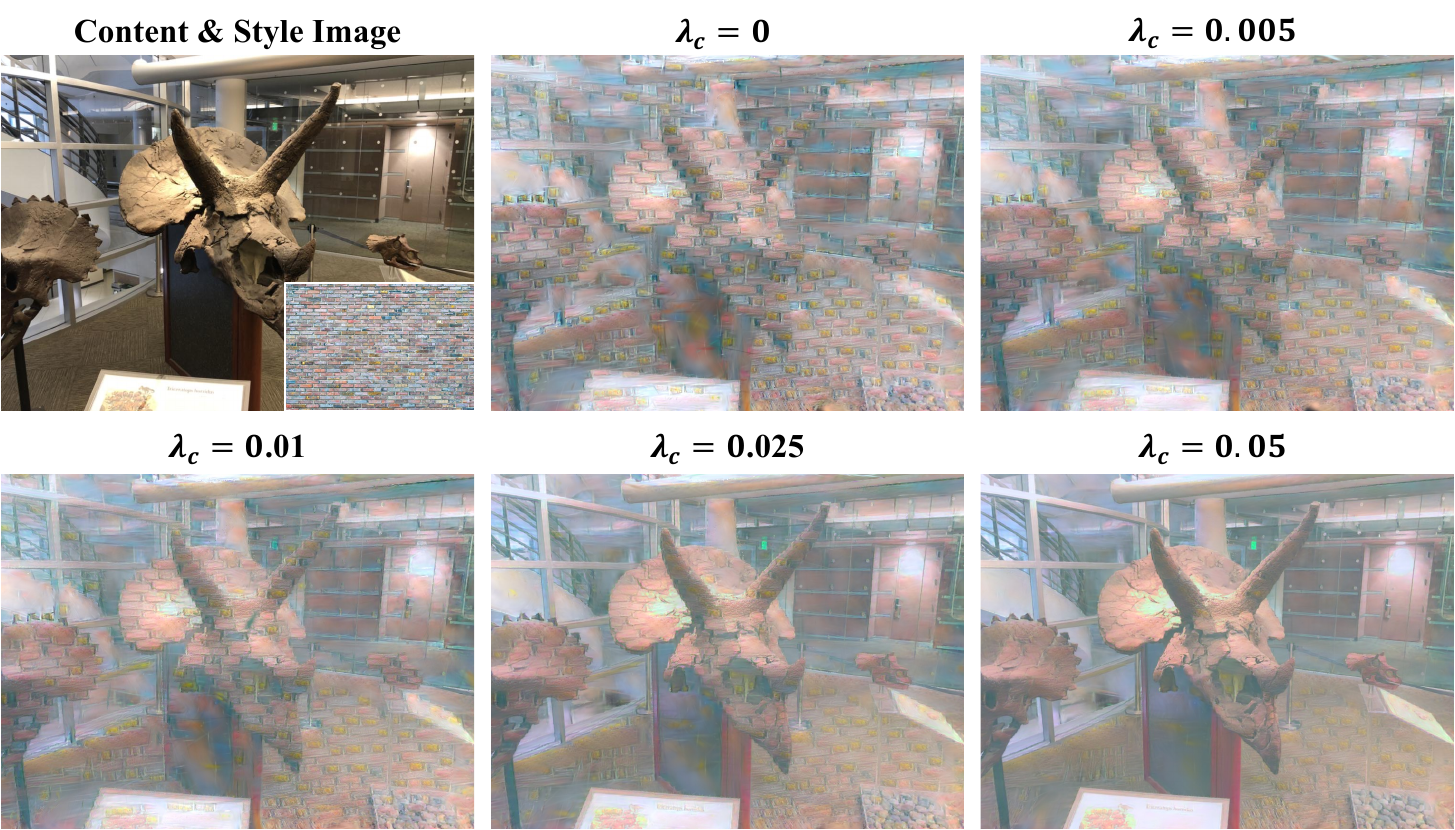}
   
   \caption{\textbf{{Ablation Studies on different $\lambda_{c}$.}} {$\lambda_{c}$ values of 0, 0.005, 0.01, 0.025, and 0.05.}} 
   \label{fig:ablation_content}
\end{figure}

\paragraph{Effectiveness of Parameter $\lambda_{c}$}
In the Stylization Stage, $\lambda_{c}$ is used to preserve content information. As shown in Fig.\ref{fig:ablation_content}, the excessively small value for $\lambda_{c}$ leads to a significant loss of content information, and the overly large value for $\lambda_{c}$ retains excessive content information, thereby suppressing the stylization effect.

\subsection{Additional Qualitative Results}
Fig.\ref{fig:single}, Fig.\ref{fig:multi2}, Fig.\ref{fig:multi3}, and Fig.\ref{fig:semantic} present additional qualitative results among different style transfer types of single-image, compositional, and semantic-aware style transfer.

\subsection{Limitations}
For limitation, like most 3DGS-based work, our
framework was sensitive to hyperparameters, and parame-
ters sometimes needed to be adjusted to achieve the desired
stylization effect. Meanwhile, since our stylization is based on the original Gaussian scene, the quality of the original Gaussian scene has a significant impact on the stylization quality.

\begin{figure*}[t]
  \centering

   \includegraphics[width=1\linewidth]{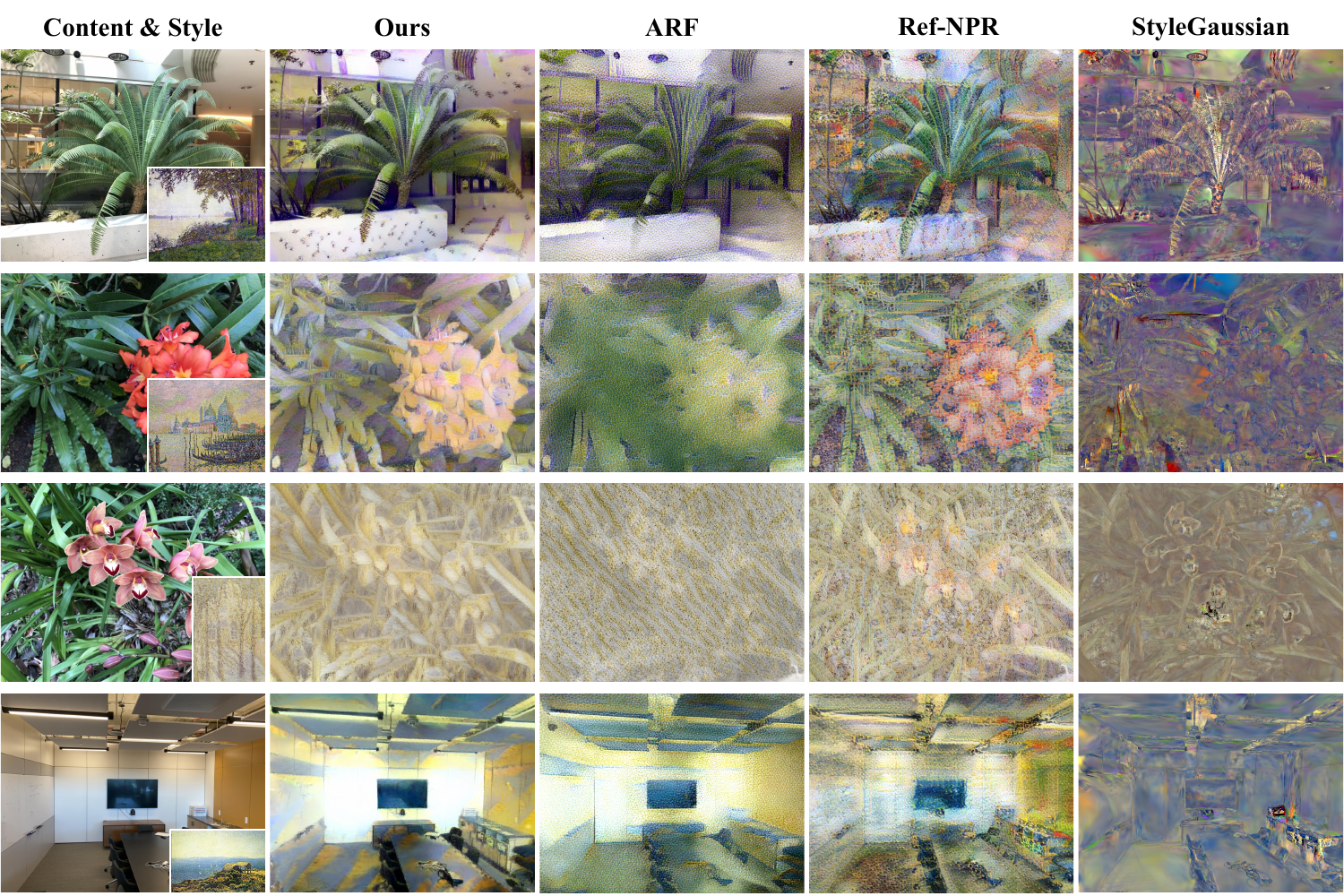}
   
   \caption{\textbf{Single-image Style Transfer.} We compare our method with the state-of-the-art methods.} 
   \label{fig:single}
\end{figure*}

\begin{figure*}[t]
  \centering

   \includegraphics[width=1\linewidth]{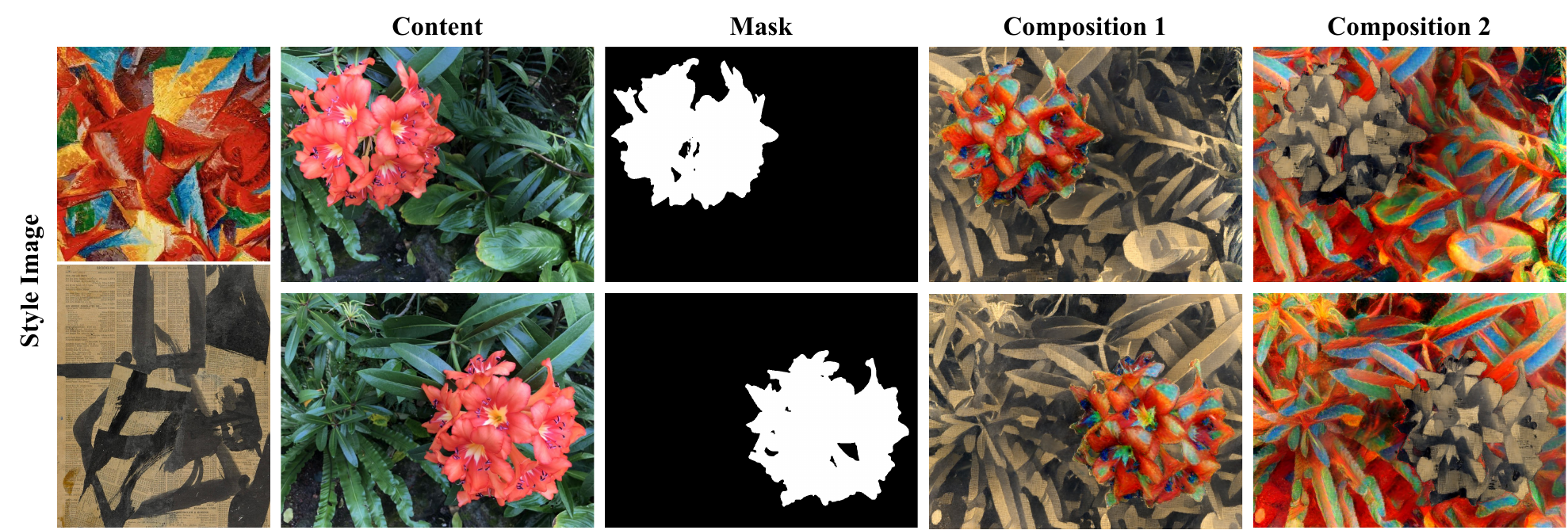}
   
   \caption{\textbf{Compositional Style Transfer.} The style is derived from two style images.} 
   \label{fig:multi2}
\end{figure*}
\begin{figure*}[t]
  \centering

   \includegraphics[width=1\linewidth]{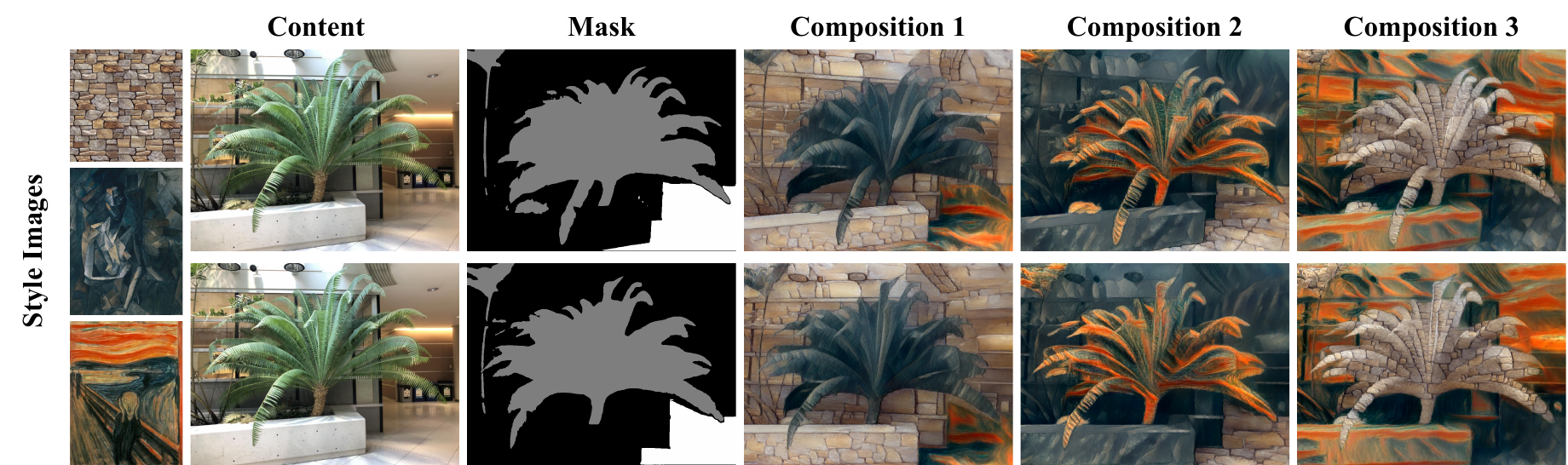}
   
   \caption{\textbf{Compositional Style Transfer.} The style is derived from three style images.} 
   \label{fig:multi3}
\end{figure*}
\begin{figure*}[t]
  \centering

   \includegraphics[width=1\linewidth]{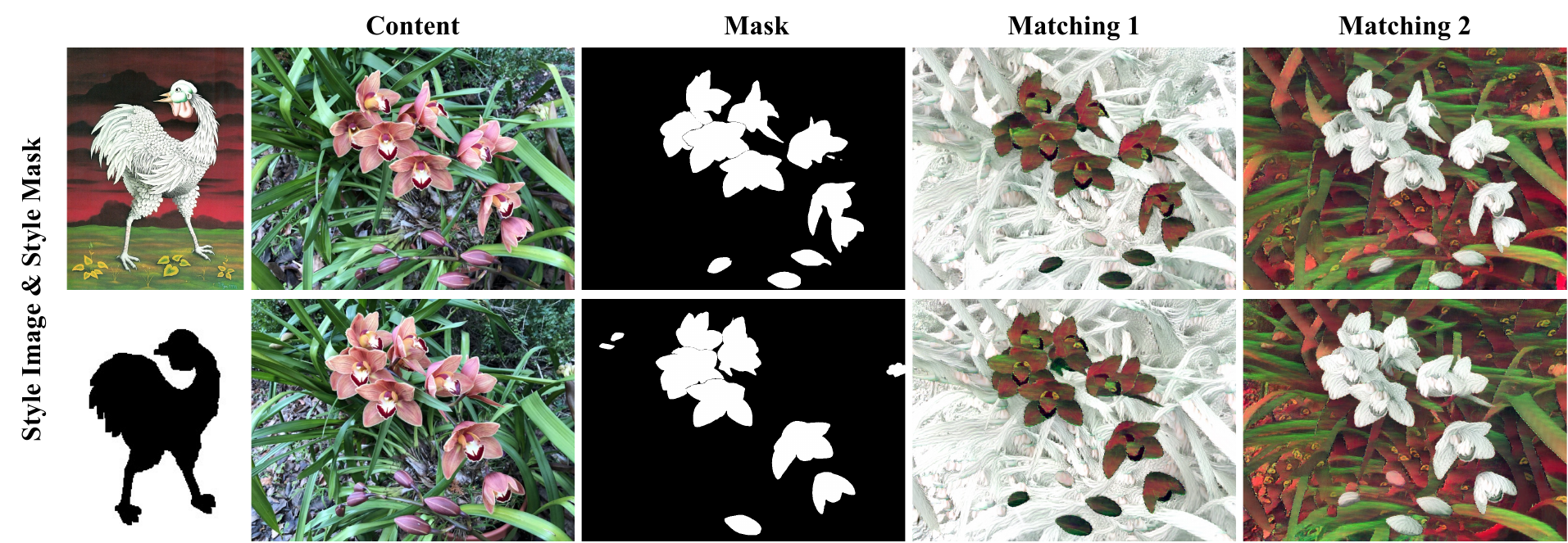}
   
   \caption{\textbf{Semantic-aware Style Transfer.} We demonstrate the stylization results achieved by matching relationships between different semantic areas.} 
   \label{fig:semantic}
\end{figure*}

\end{document}